# Learn to Solve Vehicle Routing Problems ASAP:
# A Neural Optimization Approach for Time-Constrained Vehicle Routing Problems with Finite Vehicle Fleet

Elija Deineko[a*], Carina Kehrt[a]

[a]German Aerospace Center (DLR), Institute of Transport Research, Rudower Chaussee 7, 12489 Berlin
*Correspondence: Elija.deineko@dlr.de; Tel.: +49 (30) 67055253

**Abstract**

Finding a feasible and prompt solution to the Vehicle Routing Problem (VRP) is a prerequisite for efficient freight transportation, seamless logistics, and sustainable mobility. Traditional optimization methods reach their limits when confronted with the real-world complexity of VRPs, which involve numerous constraints and objectives. Recently, the ability of generative Artificial Intelligence (AI) to solve combinatorial tasks, known as Neural Combinatorial Optimization (NCO), demonstrated promising results, offering new perspectives.
In this study, we propose an NCO approach to solve a time-constrained capacitated VRP with a finite vehicle fleet size. By means of our method, the customer-related constraints can be integrated on-the-fly, which makes it particularly flexible and transferable to numerous scenarios. We introduce an Attention Score Amplifier for Pointer Network (ASAP), a method that dynamically prioritizes urgent services during tour generation through the rescaling of attention scores in the decoder. For the vehicle fleet-related constraints, we tune the reward function with a penalty to restrict the number of vehicles. The approach is based on an encoder-decoder architecture, formulated in line with the Policy Optimization with Multiple Optima (POMO) protocol and trained via a Proximal Policy Optimization (PPO) algorithm. We successfully trained the policy with multiple objectives (minimizing the total distance while maximizing vehicle utilization) and evaluated it on medium and large instances, benchmarking it against state-of-the-art heuristics. The method is able to find adequate and cost-efficient solutions, showing both flexibility and robust generalization. Finally, we provide a critical analysis of the solution generated by NCO and discuss the challenges and opportunities of this new branch of intelligent learning algorithms emerging in optimization science, focusing on freight transportation.

**Author Contributions:** Conceptualization, E.D.; methodology, E.D.; software, E.D.; validation, E.D. and C.K.; formal analysis, E.D.; investigation, E.D.; resources, C.K.; data curation, E.D.; writing—original draft preparation, E.D.; writing—review and editing, E.D. and C.K.; visualisation, E.D.; supervision, C.K.; project administration, C.K.; funding acquisition, C.K. All authors have read and agreed to the published version of the manuscript.

**Nomenclature**

| | |
|---|---|
| AI | Artificial Intelligence |
| ASAP | Attention Score Amplifier for Pointer Network |
| BN | Batch Normalization |
| CVRP | Capacitated Vehicle Routing Problem |
| DAR | Distance-aware Attention Reshaping |
| DRL | Deep Reinforcement Learning |
| HCVRP | Heterogeneous Capacitated Vehicle Routing Problem |
| LL | Linear Layer |
| LNN | Linear Neural Network |
| MDP | Markov Decision Problem |
| MHA | Multi-Head Attention |
| ML | Machine Learning |
| MLP | Multi-Layer Perceptron |
| NCO | Neural Combinatorial Optimization |
| OR | Operation Research |
| POMO | Policy Optimization with Multiple Optima |
| PPO | Proximal Policy Optimization |
| ReLU | Rectified Linear Unit |
| RL | Reinforcement Learning |
| RLOR | Reinforcement Learning for Operation Research |
| RNN | Recurrent Neural Network |
| TSP | Traveling Salesman Problem |
| VRP | Vehicle Routing Problem |

## 1. Introduction

Generating fast and feasible solutions for complex optimization problems is increasingly important in the era of intelligent machines. The Vehicle Routing Problem (VRP) is one of the most practical and well-studied combinatorial problems in optimization science. However, the dynamic environment, numerous constraints, and stochasticities make the use of numerical combinatorial methods for real-time decision-making almost obsolete. Furthermore, the development of fast planning algorithms and decision support systems that can effectively incorporate context and handle uncertainties is crucial in the field of transportation (Heggen et al. 2019; Barua et al. 2020). Recently, a new paradigm in optimization science has emerged: Neural Combinatorial Optimization (NCO). The combination of Reinforcement Learning (RL) and generative deep learning models offers a new, vital perspective for solving any kind of sequential decision problem, including the VRP. In practice, however, the formulation of the VRP is subject to many constraints. A typical logistics service provider may be constrained by a limited number of resources, such as vehicles or drivers. In addition, it may be more beneficial to retain loads today and consolidate them in order to achieve better utilization of the vehicle fleet tomorrow. Furthermore, a typical logistics service provider is often required to serve its customers within certain time windows, e.g., by the end of the day. Thus, the VRP becomes more practical with a finite fleet and constrained arrival times. For this reason, this study deals with the Capacitated Vehicle Routing Problem (CVRP) with a finite homogeneous vehicle fleet and hard time constraints for customer services. In essence, we integrate the vehicle fleet constraints by assigning a penalty in the reward function for empty returns, i.e., if the vehicle returns loaded to the depot. The hard time constraints are implemented as a scaling factor of the attention score, effectively amplifying the attention paid to urgent services and masking the nodes where the vehicles could not deliver the service in time. In this context, the main contributions of this research are as follows:

- We extend a well-tested NCO framework for combinatorial optimization problems – Reinforcement Learning for Operation Research (RLOR) (Wan et al. 2023), which adopts OpenAI gym environment (Brockman et al. 2016) and Proximal Policy Optimization (PPO) algorithm for training (Schulman et al. 2017; Huang et al. 2022). We then extend and apply these frameworks to our problem formulation with limited vehicle fleet and hard time constraints using POMO exploration scheme (Kwon et al. 2020).
- We design and calibrate the penalty function to address the multi-objective problem i.e., maximizing vehicle fleet utilization while minimizing total fleet mileage.
- An Attention Scorer Amplifier for Pointer Network (ASAP) method for node selection with time-critical services is introduced, which can integrate hard and soft time constraints on the fly, i.e., by rescaling the attention values directly in the tour construction phase in the decoder.
- We thoroughly examine the solutions and benchmark the proposed method against the state-of-the-art heuristics.

In Section 2, the State of the Art will be presented with a focus on recent advances in NCO and related framework designs. The problem formulation together with the environmental characteristics, reward function and constraints will be explained in Sections 3.1 and 3.2. In Sections 3.3, 3.4, and 3.5 the methodology and its implementation—in particular the environment properties, encoder-decoder architecture, and the RL training algorithm—will be described in depth. The results will be shown and discussed in Section 4. In Section 5, a final summary of the research conducted and an outlook for further research will be provided.

## 2. State of the Art

In the last decade, researchers in the fields of stochastic optimization and machine learning (ML) have applied different strategies for solving dynamic optimization problems. One research group focuses on mathematical modeling, qualitatively designing Markovian models for stepwise decision-making (Simao et al. 2009; Novoa and Storer 2009;

Powell 2019, 2021; Stimpson and Ganesan 2015; Goodson et al. 2013). The research fields ML and artificial intelligence (AI) adopt data-driven approaches, designing sophisticated neural network models and tuning them with RL algorithms to find optimal solutions. Within these fields, two main approaches for solving optimization problems have emerged: hybrid methods and end-to-end NCO. Hybrid methods enhance conventional heuristics with neural optimization techniques (Hottung & Tierney 2019; Lu et al. 2020; Da Costa et al. 2020). In contrast, end-to-end NCO methods construct solutions step-by-step using encoder-decoder policy networks. This work methodologically contributes to the field of the data-driven, end-to-end solutions. Vinyals et al. (2015) introduce a pointer network to solve the travelling salesman problem (TSP) by assigning importance to each node using an attention mechanism. Nazari et al. (2017) and Bello et al. (2017) further enhance the pointer network by incorporating a simple embedding mechanism and extending the framework with RL, solving more complex combinatorial optimization problems like VRP. Joshi et al. (2019) employ a graph convolutional network to predict the sequence of nodes in a TSP. The self-attention-based encoder-decoder (also known as Transformers), introduced by Vaswani et al. (2017), is a game changer in sequential problem-solving, replacing classical recurrent neural networks (RNNs) in various domains, including combinatorial optimization. Deudon et al. (2018) generalize transformers to TSPs, achieving better results than conventional linear programming solutions. Following this methodological path, Kool et al. (2019) extend the end-to-end NCO and demonstrate its ability to solve complex combinatorial tasks. In particular, by separately encoding the depot and the service nodes, and introducing the baseline policy during the training phase, the authors achieve breakthrough results. Subsequent research has increasingly relied on this paradigm, adapting and extending it to various related variants of routing problems (Wang and Tang 2021; Bogyrbayeva et al. 2022).

However, in real-world scenarios, the VRP is subject to many constraints, such as time windows (Jungwirth et al. 2020; Homberger and Gehring 1999; Bräysy and Gendreau 2005; Gong et al. 2011), and heterogeneous or homogeneous vehicle fleets (Kritzinger et al. 2017; Eshragh et al. 2020; Ahmed et al. 2023). This poses particular challenges for NCO methods, where no established techniques for implementing optimization constraints exist. In practice, masking can be used to prevent the RL agent from performing forbidden actions (such as visiting the same node twice in VRP), leading to more efficient learning (Krasowski 2023; Spieker 2024). Through a carefully designed and calibrated reward function, actions can be manipulated by penalizing or rewarding the agent (Solozabal 2021). In the context of constrained VRPs, Falkner and Schmidt-Thieme (2020) develop an advanced attention model for VRPs with time windows, performing embeddings of both nodes and routes using an encoder-decoder attention architecture. Li et al. (2021a) exploit a similar architecture by encoding static and dynamic features for the Covering Travelling Salesman Problem (Covering TSP), i.e., the probability of selecting the next node depends on whether it can be visited directly or covered by the already visited nodes. The authors introduce the concept of a guidance vector in the embeddings and add an additional attention factor in the decoder, which together dynamically affect the probability of visiting the next node based on the nodes in the neighborhood. Li et al. (2021b) propose an attention-based Deep Reinforcement Learning (DRL) method to solve the heterogeneous VRP with vehicles of different capacities

(HCVRP). The method utilizes double decoder transformers for two sequential selection processes: to select vehicles, and to select nodes based on the selected vehicle. Peng et al. (2020) dynamically embed subproblem instances after each tour, when the vehicle returns to the depot, while constructing the VRP solution. Wang et al. (2024) demonstrate that through a direct intervention in the decoder's network, the attention score can be dynamically amplified by information from the environment. This approach forces the agent to pay more attention to the nearest nodes, and mitigate the dispersion of attentional scores in large-scale problems on-the-fly, during the learning and inference phases. Chen et al. (2024) utilize similar concept, enhancing the attention score using the distance matrix. Unlike Wang et al. (2024), they combine the product of queries and keys with the distance matrix by streaming them through an additional feedforward network.

Xin et al. (2021) point out two shortcomings in the existing NCO methods that result in reduced solution quality in the current state-of-the-art NCO methods: (I) the need for better diversity of partial solutions and (II) the use of static embeddings of the entire graph at the beginning of each episode. To mitigate the first problem, the Policy Optimization with Multiple Optima (POMO), proposed by Kwon et al. (2020), provides an effective paradigm for exploring different node sequences in the same episode in parallel, thus exploring all possible partial solutions and increasing partial-solution diversity. The POMO framework relies on trajectories[*], where each $traj \in T$ contains the partial dynamic representation of the episode at time step $t$, such as the current vehicle load, dynamical node features or other episode-specific information. Rabecq and Chevrier (2022) follow the POMO framework to handle the pick-up and delivery VRP with time windows. The authors embed multiple constraints and context information in the decoder. Fitzpatrick et al. (2023) further evaluate the performance and transferability of POMO for combinatorial optimization problems, demonstrating good solution quality on many VRP benchmark instances. Wan et al. (2023) adapt POMO approach to RL platforms and present a modular framework for solving combinatorial optimization problems–RLOR. The POMO method, in combination with the OpenAI Gym environment and RL platforms, such as CleanRL or RLLib (Huang et al. 2022a; Liang et al. 2018), shows high transferability, allowing simultaneous training and policy inference from several different nodes (i.e., learning distinctive trajectories for each time step).

The current State of the Art mostly handles VRP constraints through additional dynamic embeddings or extended decoders, which together limit the transferability and generalization of these methods. Moreover, to enhance the diversity of partial solutions, more recent studies are adopting POMO framework for combinatorial optimization. Our work integrates these two concepts by dynamically rescaling the attention score using environmental information in the decoder and by relying on the RLOR framework with POMO for training and parallel solution search.

---

[*] Note that the concept of trajectories in POMO, as denoted by Kwon et al. (2020), differs from its usage in the conventional RL context, where it refers to the tuple of states, actions, and rewards sampled by the RL agent.

## 3. Methodology

In the following section, the problem formulation will be presented and the main methodological steps, including the RL environment, the training algorithm and the policy network architecture will be described. Afterwards, the reward function calibration and the ASAP method will be shown.

*3.1. Problem formulation*

This study addresses a CVRP with homogeneous, finite vehicle fleet and predefined closing times for the customers. The problem can be formulated as follows: Given a finite fleet of vehicles with equal capacity, the objective is to serve the given demand without violating time and fleet constraints, while minimizing the total vehicle mileage. A vehicle starts the tour fully loaded at the depot and returns to the depot for reloading if it cannot reach the next customer. The customers are considered as not reachable if their demand exceeds the vehicle's current load or the transport time exceeds the defined service time limit. When the vehicle returns to the depot, the total number of vehicles in the fleet is reduced by 1. The time windows are given between 0, which means that all customers are available immediately after the first vehicle starts, and the given closing time, which represents the hard constraint for serving this customer. In fact, we have a multi-objective problem. The objectives are (i) to maximize the vehicle's utilization and satisfy as much demand as possible, and (ii) to minimize the total vehicle mileage, while not violating the hard time constraints of the customers. Such optimization problems commonly arise in scenarios such as:

- Parcel distribution in urban logistics, where carriers are not required to deliver all parcels in a single day, but can retain undelivered parcels for delivery on the following day.
- Food distribution, where grocery stores require deliveries without constrained time windows, but instead operate within standard business days for delivery.
- Customer service route planning that requires efficient routing by a company with an established fleet to provide services within specified business hours.

In the following, we will describe the extended problem formulation as well as the design of the environment, the trained policy and the training framework.

*3.2. Environment properties*

Each of the vehicles $V = \{v_1, v_2, \ldots, v_i\}$ must deliver items to $N$ customers (defined as nodes interchangeably). The set of customers $N = \{n_0, n_1, n_2, \ldots, n_i\}$ can be divided into visited customers $N^v = \{n_0^v, n_1^v, n_2^v, \ldots, n_i^v\}$ if they were visited by the vehicle $v_i \in V$, and unvisited customers $N^u = \{n_0^u, n_1^u, n_2^u, \ldots, n_i^u\}$, such that $N = N^v \cup N^u$ with $n_0$ denoting the depot. The objective is to minimize the total route length for all vehicles, serving as much demand as possible within the given resources. Each $n \in \mathbb{R}^4$, is denoted with 2-dim location coordinates (x,y), demand $d^n$ and hard time constraints i.e., end-times $e^n$. At the beginning of each episode, at time step $t = 0$, we assume that the number of vehicles $V$ is given, each with the same maximal capacity $C^v$. The current load of the vehicle $v_i$ at the time step $t$, which is used interchangeably with the vehicle utilization or its remaining capacity throughout this text, is expressed as $l_t^v$. When the vehicle's load is zero, it returns to the depot and a new vehicle is deployed, so the number of maximum possible tours is limited by $V$. It should be mentioned that this problem formulation is dynamic i.e., as time progresses during the tour, travel time of the vehicle increases. Consequently, this reduces the available time window open to fulfil the customer service. We address this by decrementing the end-time by the travel time at each time step (see Section 3.2). Without loss of generality, we assume that the total number of vehicles $V$ is always $V * l_{t=0}^v < D_{total}$. One can call this problem formulation as a prize-collecting CVRP with time windows (Balas et al. 1989; Stenger et al., 2013), whereby we do not rely on the prizes for visiting customers, rather on penalty in the reward function for the remaining vehicle load. In the following, we formulate the problem as a Markov Decision Process (MDP), which is typically defined as a tuple of state $s$, action $a$, reward $r$ and transition $Tr$: $MDP = (s, a, r, Tr)$.

**State S**: The environment state is characterized by the problem state $s_P \in \mathbb{R}^{N \times F}$, mask $m \in \mathbb{Z}^{traj \times N}$ and an information state $s_i \in \mathbb{R}^{traj}$, i.e., $S = s_p, m, s_i$, where $N$ expresses the number of nodes, $F$ is the number of node features and $traj$ is the number of given state trajectories according to the POMO definition (Kwon et al. 2020). $s_P$ is the initial, stationary state randomly generated for each instance. It contains 2-dim coordinates, demand $d^n$ and the end-times $e^n$, hence $F = 4$. The mask $m$ consists of binary variables for each node and is altered in each time step $t$ and in each trajectory $traj$, such that $m^{traj} \in \{0,1\}$. The information state $s_i$ is dynamic and contains information, such as the last node index $n_{t-1}$, the number of vehicles $V = \{v_1, v_2, \dots, v_i\}$, the current vehicle load $l_t^v$ and the already traveled distance for the entire vehicle fleet $td_t^V$. That is, for each state trajectory $s_i^{traj} = \{n_t, l_t^v, V_t, td_t^V, dones\}$.

**Action A**: The objective in our problem formulation is to construct tours, i.e., a sequence of visited nodes for each vehicle $v_i$, such that $N^{v_i} = \{n_0^{v_i}, n_1^{v_i}, n_2^{v_i}, \dots, n_i^{v_i}\}$. The action space is therefore the indexes of the visited nodes for each trajectory $a^{traj} = \{n_t^v\}$. In this context, an action in step $t$ is the next node to visit by the vehicle $v^{traj}$. If the next node demand exceeds the current vehicle capacity, or the node is not reachable because the end-time expires (i.e., $e^n \leq 0$), or all nodes are masked, the vehicle returns to the depot and $a^{traj} = n_0^v$.

**Transition**: The transition function defines the update rules for the current state $s_t$, after taking the action $a_t$, i.e., when the environment transitions into a new state $s_{t+1}$. In each state trajectory, the current state components become updated by the following rules.

$$m_{t+1}^n = \begin{cases} 0, \forall n^u \text{ if } t = 0 \text{ and } \forall n_0 \text{ if } t > 2 \\ 1, \forall n^v \text{ if } n_{t+1} = a_t, \text{ or } d_{t+1}^{n^v} > l_t^v, \text{ or } l_t^v = 0, \text{ or } V = 0 \end{cases} \quad (1)$$

$$l_{t+1} = \begin{cases} l_t^v - d_{t+1}, \forall n^v \\ l_t, \quad \text{otherwise} \end{cases} \quad (2)$$

$$v_{t+1} = \begin{cases} V - 1, \text{ if } a_t = 0 \\ V, \quad \text{otherwise} \end{cases} \quad (3)$$

$$td_{t+1}^V = td_t^V + dist(n_t, n_{t+1}) \quad (4)$$

$$e_{t+1}^n = e^n - td_t^V / S^v \quad (5)$$

Equation 1 expresses, that in initial state $t = 0$ the mask $m_{t=0}^n$ contains zeroes for all unvisited nodes $n^u$, except the depot $n_0$, but the depot is unmasked after two steps i.e., $t > 2$. We switch the visited nodes in mask to 1, if the node was visited, i.e., $n^v = a_t$. Additionally, we explicitly mask all nodes, where the demand exceeds the current vehicle load: $d_{t+1}^n > l_t^v$, or if the remaining load of the vehicle is zero: $l_t^v = 0$, or no more vehicles are available: $V = 0$. If the node is not reached and the depot is not masked ($m_t^{n_0} = 0$), the vehicle returns to the depot and $n_0$ becomes $m_{t+1}^{n_0} = 1$.

Furthermore, the load of the current vehicle is reduced by the demand of the visited customer in Equation 2. Equation 3 denotes that the number of the vehicles is reduced in each state, if the last action is a depot: $a_t = 0$. Equation 4 is just the cumulative distance traveled by the vehicle fleet $V$. Equation 5 defines the reduction of the service end-times $e^n$ for each customer by the cumulative traveled time of the entire vehicle fleet $td_t^V / S^v$. Finally, after transiting in $s_{t+1}$, the last node becomes the current node: $n_{t+1} = a_t$. The problem is terminated either when all vehicles are deployed, i.e., $V_{t+1} = 0$, or there are no customers to visit $m_{t+1}^{n=N} = 1$.

*Reward:* To encourage the agent to visit as many customers as possible (because otherwise the agent will always select the depot in the next step to reduce the total distance), we formulate the reward function as follows:

$$r_t = min(dist(n_t, n_{t+1}^u) + P * l_t^v) \text{ for } t \leq N, \quad (5)$$

where *dist* is a Euclidean distance between the next unvisited node $n_{t+1}^u$ and this node $n_t$. $l_t^v$ is the load of the vehicle $v$ at the time step $t$. The penalty factor $P$ penalizes the agent to return to the depot with the full load $l \to 1$ and does not, when the load of the vehicle $l \to 0$. In order to adjust the penalty signal and avoid noise in the reward function, we need to tune the penalty factor $P$ with respect to other components of the reward function. If the penalty for not serving the customer is too small, the agent will choose to return to the depot as soon as possible and will not use the full capacity of the vehicle. If the penalty signal is too large, the reward function will be too noisy and the agent will only seek to exploit the full load at all costs (Tessler et al. 2018; Solozabal et al. 2023). In the following, the choice for the penalty factor will be explained and graphically shown.

Without loss of generality, we assume that the average distance between two random nodes $i$ and $j$ is 0.5 If the load $l_{t=0}^v = 1$ and the average customer demand $d^n \sim 0.125$, the maximum tour size for the vehicle will approximately be 8 customers, or $n^v \sim 8$ and the maximum mileage for vehicle $v$: $dist_v = 4$. In order to assign a proper negative penalty to the agent if he returns to the depot too quickly, we penalize the agent by approximately $2.5*dist_v$. Figure 1 depicts how the environmental components such as rewards, vehicle loads, number of vehicles and the tour length evolve during the route construction phase with a penalty factor $P = 10$.

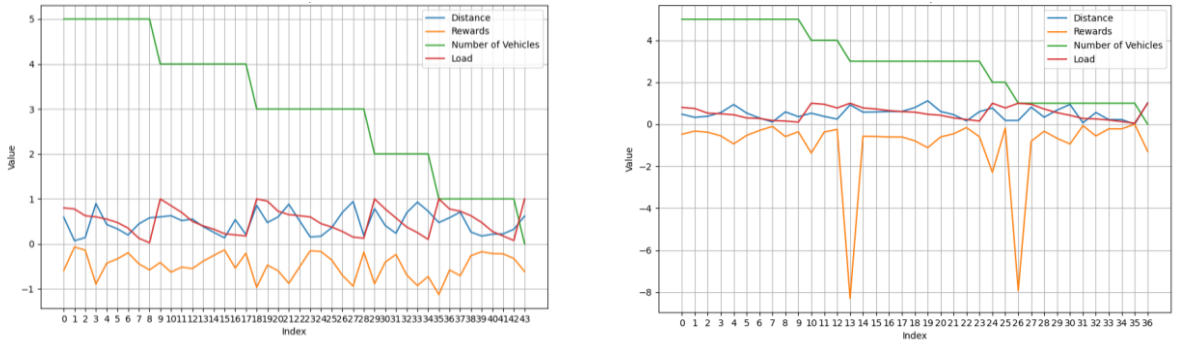

Figure 1: Parameters from the environment such as distance (blue line), rewards (orange line), number of vehicles (green line) and the current load (red line) during two selected episode rollouts. The decrement in the number of vehicles (green line) indicates the return to the depot after $t$ steps. The left figure represents the episode where the vehicle's load is fully exploited in each tour. The right figure contains tours with not utilised vehicle load. The number of steps $t$, i.e., the number of nodes served by the vehicle, are depicted on the x-axis.

As shown in Figure 1, with $P = 10$ the reward $r$ varies between 0 and 1 for long tours (Figure 1, left), when the vehicle load is fully exploited. The reward decreases up to -8, when the vehicle returns to the depot too early, after serving only several customers (Figure 1, right). Hence, we conclude that the penalty decreases in the feasible proportion to the distance travelled. We tried several penalty factors, but $P = 10$ worked the best for the problem with $N = 50$.

### 3.3. Attention Scorer Amplifier for Pointer Network (ASAP) method

Inspired by the Distance-aware Attention Reshaping (DAR) method introduced in Wang et al. (2024), we implement our customer time constraints directly in the decoder by rescaling the attention score in the pointer network by adding an *urgency factor*. In Wang et al. (2024), the attention values in the decoder are adjusted based on the node proximity, resulting in the enhanced attention to the nearby nodes within a tour. We use similar logic, but integrate time constraints as the attention rescaling factor, which we call the *urgency factor* for the customers. A more detailed description for this procedure will be provided in the following. The urgency factor $f_u$ is calculated as follows.

$$f_u^{n_t} = \frac{dist(n_t, N^u)/v_s}{e^n - td_t^V/S^v} \qquad 0 < f_u < 1, \in \mathbb{R}^{traj \times N} \tag{6}$$

$f_u$ expresses the urgency for serving all unvisited nodes $N^u$ for the vehicle being in the current node $n_t$. In other words, it computes a ratio between anticipated travel times and the end-times, which are updated in each time-step $t$. First, we determine the distance between the actual node $n_t$ to all unvisited nodes $N^u$ in this episode and in each trajectory as: $dist(n_t, N^u)$. We divide these distances by the updated end-times for each node: $e^n - td_t^V$, where $e^n$ are given end-times for each customer and $td_t^V/S^v$ is the total time travelled so far by the vehicle fleet V, updated in each time step $t$ and each trajectory. The factor $S^v$ represents the vehicle velocity and is needed to convert the distances into travel times. Consequently, we assume that $S^v = 50\frac{km}{h} = 0.014 \frac{distance\ units}{s}$ for the normalized instances with $x, y \in [0,1]$. Therefore, if $f_u \to 0$, then the end-time requested by the customer lies ahead and there is no need to urge. If $f_u \to 1$, the service becomes time critical. For $f_u > 1$ or $f_u < 0$ the service is considered as missed, and we set this customer to $m_t^n = 1\ \forall\ n$ where $0 < f_u < 1$, effectively masking the node to be selected. For further implementation details of the ASAP method see Section 3.4, Equation 19.

*3.4. Policy network*

The implementation of our policy network is based on Wan et al. (2023), which adapts the well-tested model by Kool et al. (2018) and extends it to the trainable framework CleanRL with POMO. CleanRL is an algorithm library that provides implementations of a wide range of state-of-the-art RL algorithms for policy training, hyperparameter tuning, and experimental results tracking (Huang et al. 2022a). However, the policy training process in the standard CleanRL implementation differs from the original end-to-end NCO pipeline in many aspects. First, most RL training platforms usually cannot train only a part of the policy network (e.g., decoder), forcing a forward pass through the entire encoder-decoder networks at each step of the episode (see Wan et al. 2023). Second, in the original RL implementation by Kool et al. (2018), the loss in the REINFORCE algorithm is calculated as $L(\pi|s) - b(s)$, where $L(\pi|s)$ represents the return, and $b(s)$ is the baseline, often defined as the 'best policy so far'. In CleanRL, however, the value of the current state predicted by the critic network is used as a baseline to reduce loss variation. We will now present the core principles of our policy network implementation and provide a comprehensive description of the refactoring steps needed to achieve multi-constrained policy training for the proposed problem.

The policy network design implemented by Wan et al. (2023) is an open-source encoder-decoder model with an 8-head self-attention network ($MHA$). First, the problem state $s_P$ is embedded through a linear neural network (LNN) $emb_n = nW_l^{dim} + b^{dim}$ with $n = \{n_1, n_2, n_i ...\}$ and $emb_{n_0} = n_0 W_l^{dim} + b^{dim}$. Here, the size of the network is determined by the weight matrix $W$ of dimensions dim=128. By concatenating the depot and node embeddings, we obtain an embedded representation for the problem state across each batch $B$: $emb_{s_p} \in \mathbb{R}^{B \times N \times 128}$. Note that we specifically embed only the static components of the problem state $s_P \in \mathbb{R}^{N \times 3}$, which include node's coordinates and demand values, and are independent of the trajectories $traj$ (see Figure 2).

**Encoder:** The encoder $MHA$ contains 3 layers $l^{1,2,3}$, which consume the embedding tensors $emb_{s_p}$ and calculates queries $Q$, keys $K$ and values $V$ as linear projections:

$$Q_l = emb_l W_l^Q, \qquad K_l = emb_l W_l^K, \qquad V_l = emb_l W_l^V, \tag{6}$$

with $Q_l, K_l, V_l \in \mathbb{R}^{B \times N \times 128}$ and $W_l^Q, W_l^K, W_l^V$ being trainable parameters for each layer $l$. We decompose the projected $Q_l$, $K_l$, and $V_l$, into $Q_{l,H}^{1,2,3} \in \mathbb{R}^{H \times B \times N \times 16}$, $K_{l,H}^{1,2,3} \in \mathbb{R}^{H \times B \times N \times 16}$, $V_{l,H}^{1,2,3} \in \mathbb{R}^{H \times B \times N \times 16}$, where $H = 8$ is the number of heads. The attention score $u_H$ is calculated as:

$$u_H = softmax\left(\frac{Q_{l,H} K_{l,H}^T}{\sqrt{\dim_K}}\right) V_{l,H} \text{ where } u_H \text{ becomes } u_H \in \mathbb{R}^{H \times B \times N \times 16} \tag{7}$$

$$u_{emb}^{MHA} = Concat(u_H) W_l^h, \text{ resulting in } u_{emb} \in \mathbb{R}^{B \times N \times 128} \tag{8}$$

Again, $W_l^h$ denotes a trainable linear layer ($LL$) of the size $W^{128 \times 128}$. The $MHA$ output $u_{emb}^{MHA}$ is then sequentially passed through multiple normalization, linear, and activation layers for each $l$. The overall calculation procedure is summarized as follows:

$$h_l^1 = BN(u_{emb}), \tag{9}$$

$$h_l^2 = FF_l^2(ReLU(h_l^1 * LL_i^{128 \times 512}), LL_j^{512 \times 128}), \tag{10}$$

$$h_l^3 = BN(h_l^2), \tag{11}$$

$$h_{enc} = FF_{enc}(h_{l=1}^3, h_{l=2}^3, h_{l=3}^3), \tag{12}$$

where $BN$ is a Batch Normalization layer (Equation 9), $FF_l^2$ in Equation 10 is a sequential feed forward network with $LL_i$ and $LL_j$–two linear layers defined as:

$$LL_i = xW_i + b_i, \text{ where } W_i \in \mathbb{R}^{128 \times 512} \text{ and } b_i \in \mathbb{R}^{512}, \tag{13}$$

$$LL_j = xW_j + b_j, \text{ where } W_j \in \mathbb{R}^{512 \times 128} \text{ and } b_j \in \mathbb{R}^{128}. \tag{14}$$

After applying a $ReLU$ activation layer in Equation 10, and another $BN$ in Equation 11, the encoder produces the hidden state $h_{enc}$ of the size $h_{enc} \in \mathbb{R}^{B \times N \times 128}$ for each node in $N$, after processing through three $MHA$ layers $l_{1,2,3}$ (Equation 12).

**Decoder:** The decoder consumes the input from the encoder networks $h_{enc}$, along with additional state information $s_i$ (i.e., the mask and the context), and produces a probability distribution over all reachable nodes for each trajectory traj. Typically, the decoder consists of several linear layers for graph and context projections, an $MHA$ network, and an attention score function called a pointer network, which produces the final output. We build upon the architecture from Wan et al. (2023), integrating dynamical time constraints into the pointer network, which we refer to as an ASAP method (see Section 3.3). Below, the main implementation steps of the decoder are described in depth.

(1) To obtain a more complete graph representation of the initial problem instance (a global graph context vector), we first average the hidden representations $h_{enc}$ produced by the encoder in Equation 12 over all nodes N:

$$c_h = \frac{1}{N}\sum_{i=0}^{N} h_{enc}, \qquad (15)$$

resulting in $c_h \in \mathbb{R}^{B \times 128}$. This vector is then transformed through a linear layer: $c_h = c_h W_h$, with the trainable parameters $W_h \in \mathbb{R}^{128 \times 128}$.

(2) Additionally, by feeding $h_{enc}$ through a linear layer $LL_g$ of the size $W_g \in \mathbb{R}^{128 \times 384}$, the aggregated projection for nodes is calculated: $g = LL_g(h_{enc})$. Effectively, this layer expands the encoded nodes to a size of (3 ∗ dim). $g$ is then decomposed again into $g_k, g_v$ and $k_L$, each of the size $g_k, g_v, k_L \in \mathbb{R}^{B \times N \times 128}$.

(3) The problem context, i.e., the vehicle load and the number of the remained vehicles in each state trajectory, is concatenated in the context state: $c_{state} = concat(l^v, V)$, and then concatenated again with the current node embedding $emb_{n_t}$:

$$c_o = concat\left(emb_{n_t}, c_{state}\right), \text{resulting in } c_o \in \mathbb{R}^{B \times traj \times 130}. \qquad (16)$$

This step effectively combines relevant context information, the vehicle load and the vehicle fleet information, with the hidden representation of the last visited node.

(4) The resulting context $c_o$ is then passed through a linear layer again: $c_o = c_o W_o$, where $W_o \in \mathbb{R}^{130 \times 128}$, resulting in an output of a suitable size for further processing: $c_o \in \mathbb{R}^{B \times traj \times 128}$.

(5) The query $q_m$ for further attention score calculation is obtained by adding the graph context $c_h$ to the state context $c_o$ as:

$$q_m = c_h + c_o, \text{resulting in } q_m \in \mathbb{R}^{B \times traj \times 128}. \qquad (17)$$

(6) This query $q_m$, together with the keys and values $g_k$ and $g_v$ from Step 2, and the mask $m \in \mathbb{R}^{B \times traj \times N}$ from the environment are passed to another instance of the $MHA$, identical to the encoder's $MHA$ described in Equations 6–8:

$$g_Q = MHA_{dec}(q_m, g_k, g_v, m), \text{producing } g_Q \in \mathbb{R}^{B \times traj \times 128}. \qquad (18)$$

(7) Finally, to compute the probability distribution over all candidate nodes $N$ in each trajectory traj and batch B, attention score calculations are performed within the pointer network. The pointer network computes the attention factor $u_{dec}$ as the dot product between the current node and all other embedded nodes:

$$u_{dec} = \left( \tanh\left[ \frac{g_Q * k_L^T}{\sqrt{dim_{k_L}}} * m \right] + f_u \right) * C, \text{ resulting in } u_{dec} \in \mathbb{R}^{B \times traj \times N} \tag{19}$$

First, we inject additional time and distance-related information $e^n$ and $td$ into the pointer network direct from the environment and calculate the urgency factor $f_u$ as given in Equation 6. In Equation 19, $g_Q$ refers to the state query from Step 6, and $k_L$ represents the projected keys for each node from Step 2. The $\tanh(u)$ function rescales the attention scores u in the range of $-1 < u < +1$. Next, $f_u$ is added to augment the attention scores for nodes that need to be served urgently ($f_u \to 1$). If the service urgency is small ($f_u \to 0$), we simply rely on the attention values provided by the decoder. Nodes that are either masked by $m$ or unreachable, i.e., due to the expired end-time $e^n$ (if $f_u > 1$), are assigned to $-inf$. We then apply a clipping factor $C = 10$ to rescale the overall attention score in the range $-10 < u_{dec} < 10$.

(8) The next action $a$ is then sampled from the categorical probability distribution $P_{cat}$, ensuring that the agent learns the stochastic policy:

$$a = P_{cat}(u_{dec}), \text{ where } a \in \mathbb{R}^{B \times traj}. \tag{20}$$

Note, that during the training loop, we sample actions for each state trajectory from the distribution in Equation 20. For inference, however, the next node is selected based on the highest attention score: $a = max(u_{dec})$. Analogues to Wan et al. (2023), the decoder output $u_{dec}$ is then supplied to RL training algorithm. Lastly, the state value $v_{critic}$ produced by the Critic network, is then derived as follows:

$$v_{critic} = MLP_{critic}(g_Q) \tag{21}$$

$MLP_{critic}$ consists of two fully connected layers, $LL_1^c$ and $LL_2^c$, which sequentially process the input tensor $g_Q$ and output a scalar state value $v_{critic}$: $LL_1^c = g_Q W_c^{dim_{g_Q} \times 128} + b_c^{128}$ and $LL_2^c = xW_c^{128 \times 1} + b_c^1$. Here, the $\dim_{g_Q}$ denotes the dimension of the input tensor $g_Q$. The general architecture of the policy network is illustrated in Figure 2.

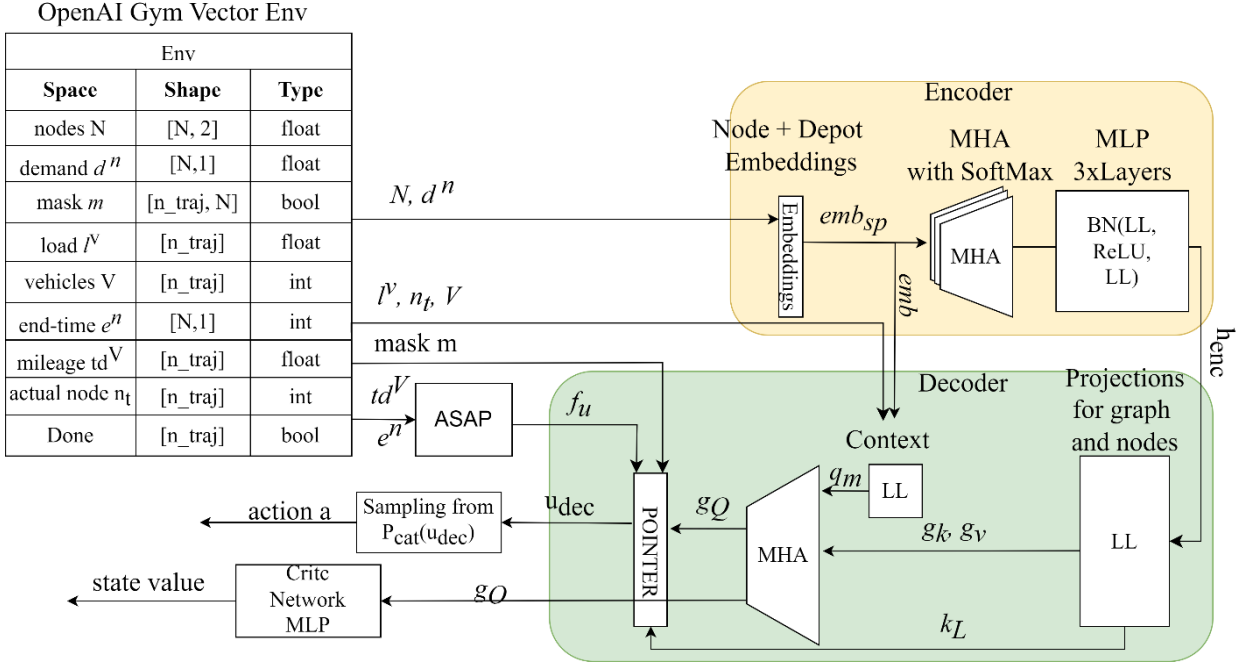

Figure 2: Policy network architecture with its input and output data flows. First, the static problem state is processed through an *MHA* encoder network with three MLP layers. Next, the hidden problem representation from the encoder $c_h$, together with the state information from the environment (load $l^v$, current node, and the number of remained vehicles $V$), is added to the context vector and supplied to the decoder. The dynamic time constraints of each node are converted into the urgency factor $f_u$ via the ASAP method, according to Equation 6, Section 3.3. This factor $f_u$ together with the mask m are then are passed to the pointer network, prioritizing nodes that require immediate service.

### 3.5. Reinforcement learning training

In the following sections, we will describe the training algorithm for the policy network in depth. We collect the batches of size $B$ from synchronized parallel vector environments, where each batch contains the state trajectories traj. At each iteration, we use these batches and their trajectories within the PPO algorithm to update the network weights. Our training procedure follows the PPO implementation described in Huang et al. (2022), which was further adopted and extended to NCO problems with POMO by Wan et al. (2023). The pseudocode with the training procedure is presented in Algorithm 1.

Algorithm 1: Implementation of the applied PPO training algorithm.

```
1  1. Rollout Phase.
2  Collect observations stepping the vectorized environments env in parallel.
3  Initialize policy π₁.
4
5  For U global updates:
6      For each step t ∈ T, where T is a given number of steps in the env, do:
7          Sample observations obs from the env, as described in Section 3.2.
```

Feed $obs$ into the policy network $\pi_1(obs)$, as described in Section 3.4.

*Return*: next action $a_{\pi 1}$, action probability $P(a_{\pi 1})$ and the next value $v_{\pi 1}$.
Bootstrap next state values by reverse computation from the end of each episode starting with the last step $t_{T-1}$ for every batch $b \in B$:

Compute Advantages using Generalized Advantage Estimation (GAE) for policy $\pi_1$ if episode is not $done$:

$A_t = \delta_t + \gamma * \lambda * A_{t-1}$, where $\gamma$ and $\lambda$ are given coefficients,
and $\delta_t = r_t + \gamma v_{\pi_1(t+1)} - v_{\pi_1(t)}$
else: $A_t = r_t - v_{\pi_1(t)}$.

Compute the return from $\pi_1$: $R_{\pi_1} = A_t + v_{\pi_1}$, where $R_{\pi_1} \in \mathbb{R}^{T \times B \times traj}$

2. Update policy network.
Flatten collected actions $a_{\pi_1}$, action probabilities $P(a_{\pi_1})$, next value $v_{\pi 1}$, $R_{\pi_1}$ and $A_{\pi_1}$.
Form minibatches of the size $b_x = \mathbb{R}^{T*\frac{B}{n} \times traj}$, where $x$ – shuffled batch indices; T defines the number of steps in the environment; $n$ is a number of given minibatches, and $B$ is a total number of batches.
 For each *epoch* update policy $\pi_2$:
  For each minibatch $b_x$ do:
   Sample $b_x$ observations from the policy $\pi_2$: $obs_{b_x, \pi 2} = \{s_{\pi 2}, a_{\pi 2}, P(a_{\pi 2}), v_{\pi 2}, e_{\pi 2}\}$

   Calculate a ratio between the old action probabilities under $\pi_1$
   and the new action probabilities under $\pi_2$ as: $L_r = \exp(P(a_{\pi_1}) - P(a_{\pi_2}))$

   Calculate policy loss: $L_p = mean(L_r * (-A_{\pi_1}))$, where $L_p$ is a mean policy loss

   Calculate value losses: $L_v = mean(0.5 * (v_{\pi 2} - v_{\pi 1})^2)$, where $L_v$ is a mean value loss

   Compute average entropy from $\pi_2$: $e_l = mean(e_{\pi 2})$

   Calculate total loss from policy losses, entropy values and value losses:
   $L_{total} = L_p - E^{coef} * e_l + L_v * V^{coef}$, where $E^{coef}$ and $V^{coef}$ are given coefficients

   Perform backpropagation and update policy $\pi_2$ using $L_{total}$.

   Assign updated policy $\pi_1 = \pi_2$ and repeat the training for $U$ timesteps

We start by sampling each episode stepping through $n$ parallel, independent environments and forming batches $B$. In the first phase of Algorithm 1, we step the environment 100 times (i.e., collecting 100 observations) in $n = 1{,}024$ parallel environments, resulting in a total of 102,400 batches with 50 trajectories generated for training. Note, that in PPO we have a fixed number of steps for each episode (T=100). In the second phase, we flatten the batch $B$ and split it into 8 minibatches $b_x \in B$, each containing 128 episodes. In each training epoch, we shuffle the samples and compute policy losses, entropy values, and value losses as outlined in Algorithm 1. Finally, the total policy loss is then used to update the policy $\pi_2$. The rollout and learning phase are iterated for approximately $U = 10{,}000$ global update steps. These and other hyperparameters, such as the learning rate $\gamma$ and the entropy coefficient $E^{coef}$, are set analogous to CleanRL protocol (see Huang et al. 2022). The policy is evaluated every 100 global update steps. For detailed information on implementation and hyperparameter selection, refer to Wan et al. (2023).

## 4. Experiments and result analysis

In the following, the proposed framework will be trained and validated through an in-depth examination of the inferred solution and benchmarking against state-of-the-art heuristics for VRP.

### 4.1. Training instances

We generate the training and evaluation problem instances for each episode, following the procedure analogous to the comparable studies in the State of the Art, such as Kool et al. (2018), Wan et al. (2023) and Berto et al. (2023). Each parallel environment is initialized with 50 customer locations and a depot, randomly distributed in the normalized Euclidean space (i.e., $0 \leq x, y \leq 1$), resulting in $N = 51$. The node features are then extended by the demand and end-times values, according to Section 3.2. Each problem instance is parametrized by the vehicle capacity $C^v = 40$, which is then normalized to $l^v_{t=0} = 1$. We further distribute the demand uniformly: $d_n \sim P_u(1,10)$ for $n = 1, 2, \ldots, N$ and normalize the demand by $d^n / C^v$. This ensures that the customer demand is always smaller than the vehicle capacity and the minimum tour length for the vehicle $v_i$ becomes $\frac{C^v}{\max d^n} \to 4$ customers, when operating at full capacity. The total number of vehicles in each instance is set to $V = 5$. Service end-times constraints $e^n$ are expressed in time units and are randomly sampled from the range $50 \leq e^{n \neq 0} \leq 10{,}000$ for customers, and $e^{n=0} = 10{,}000$ for the depot. This range defines the minimum and maximum travel times for a vehicle travelling at 50 km/h in a normalized instance. For example, a value of $e^n = 10{,}000$ is equivalent to travelling 15 times the maximum distance at a constant speed of 50 km/h (see Section 3.3). This constraint ensures that the most customers are reachable within the reasonable time span.

## 4.2. Results

To assess the quality of the decisions produced by the trained model, we initially focus on deploying a greedy policy inference on small instances and showcase the pointer output. The model is trained for 10,000 update steps on the datasets specified in Section 4.1. The inference is performed on a single instance with $N = 10$ customers and end-times $50 \leq e^n \leq 500$. According to Equation 19, the urgency factor $f_u$, defined as the ratio between the time traveled to the customers and the given end-times $e^n$, is added to the attention score in the pointer network. This effectively highlights urgent services resulting in the higher attention scores for customers with $f_u \to 1$. If the customer service end-time in time step t+1 lies far ahead (i.e., $e_{t+1}^n \sim 500$) and the distance from this node $n_t$ to the next unvisited node $n_{t+1}^u$ is short (i.e., $dist(n_t, n_{t+1}^u) \to 0$), the attention score in the pointer network remain unchanged, as postulated in Equation 6 and Equation 19. Figure 3 (left) demonstrates the solution generated by the trained ASAP-NCO model. The total mileage in this solution is 3.06 distance units. Figure 3 (right) presents the attention score distribution (logit values) for each node (y-axis) and for each step (on the x-axis).

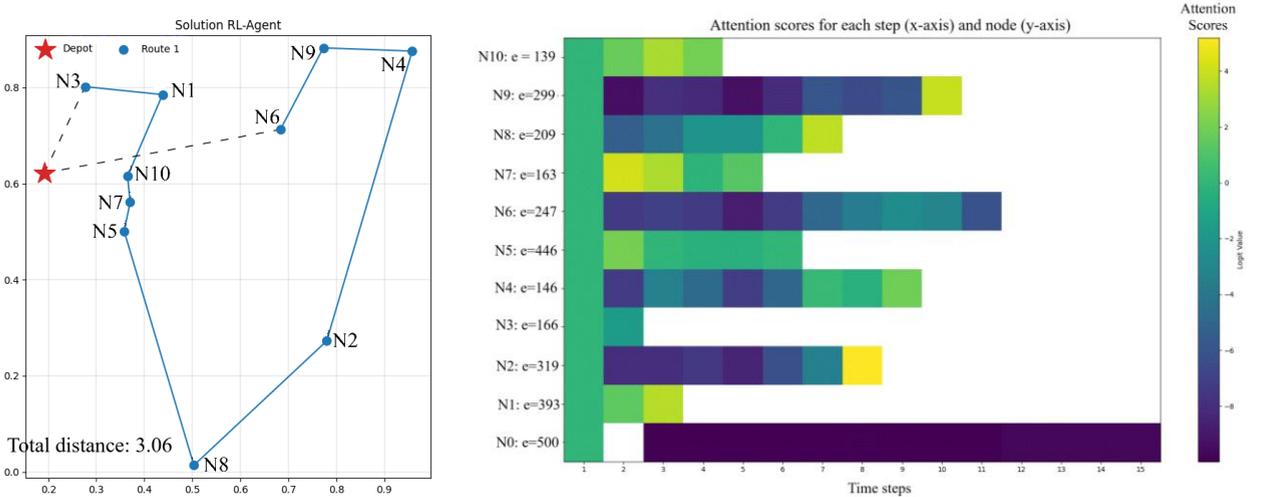

Figure 3: On the left, the tour constructed by the trained model for a small test instance with 10 nodes. The solution, including the total mileage, is shown in the figure on the left. On the right, the decoder's output, obtained according to Equation 19 for each environment step (x-axis) and each node (y-axis), is plotted. The different colors along the y-axis of the right figure represent the attention score values for each time step and node.

In the initial time step, the agent starts at the depot, and all attention scores are zero, as shown in Figure 3 (right). The first node, i.e., node 3 (Figure 3, left), has the third smallest service end-time ($e^{n=10} = 139 < e^{n=7} = 163 < e^{n=3} = 166$). Due to the shortest distance and the shortest end-time, the agent assigns the highest attention score to this node N7 (see Figure 3, right. The logit value for $e^{N7} = 163$). However, the first starting node is N3, which is the second-best option when following the greedy policy. Since we are in POMO, the agent builds all 10 trajectories in the initial steps and considers all 10 nodes as equally suitable starting points. By doing this, we effectively overcome the issue of a poor initial solution node, even though its logit was the highest in the first step. In the time step two, the agent selects node N1, probably due to the closest distance, as shown in Figure 3 (left). In step $t = 3$, the N10 exhibits the highest attention value due to its narrow end-time $e^{n=10} = 139$ and the shortest distance to N1. As illustrated in Figure 3, all remaining nodes are selected according to the greedy policy, i.e., following their highest attention scores. As a result, all 10 customers are visited without violating the time constraints, forming a reasonable and consistent tour.

*4.3. Benchmarking*

The second objective of our validation analysis is to compare our ASAP-NCO solution against the established Operation Research (OR) heuristics to examine the overall performance and generalization of the trained model. The results of our model are then benchmarked against a state-of-the-art optimization framework for multi-constrained VRP—PyVRP (Wouda et al. 2024). PyVRP efficiently combines a genetic algorithm with a local search algorithm and can be extended to various OR problems, including CVRP and VRP with time windows. We adapt this framework to our problem setup by scaling the instances by a factor of 10,000 and formulating the problem as a prize-collecting CVRP with time windows and finite vehicle fleet. For the time constraints, we simply set the customer time windows constraints to the given customer end-times $e^n$ in PyVRP. The prize collection factor, which rewards heuristics for visiting customers, is set to 1,000 for large instances (with up to 1,000 nodes) and 5,000 per customer for small instances with fewer than 500 nodes. This parameter needs to be calibrated to reflect the trade-off between minimizing the total distance and maximizing the number of nodes visited by the heuristics. As a consequence, PyVRP search heuristics for time constrained prize-collecting CVRP behaves similarly to the trained NCO agent. Table 2 shows the results in terms of solution quality for PyVRP and the trained model.

Table 2: Solutions from the benchmark heuristics PyVRP (indicated as OR solution) and the proposed ASAP model.

| Instances | | | | Solution | | |
|---|---|---|---|---|---|---|
| Index | Nodes | Seed | Prize coll. factor for OR heuristics | OR solution | NCO-ASAP (proposed model) | Relative Deviation |
| N50a | 50 | 1234 | 5,000 | 7.59 | **6.33** | +17% |
| N50b | 50 | 4321 | 5,000 | 8.68 | **8.05** | +7.3% |
| N100a | 100 | 1234 | 2,500 | 7.70 | **6.25** | +19% |
| N100b | 100 | 4321 | 2,500 | 7.71 | **6.78** | +12% |
| N500a | 500 | 1234 | 1,000 | 5.72 | **3.07** | +46% |
| N500b | 500 | 4321 | 1,000 | 4.82 | **4.22** | +12% |

| N1000a | 1,000 | 1234 | 1,000 | 7.56 | **4.48** | +40% |
| N1000c | 1,000 | 3241 | 1,000 | 6.22 | **4.67** | +25% |

For validation, eight cases were generated, either with varying distributions (seeds) or different numbers of customers. As mentioned in Section 4.1, the model was trained on random instances with 50 customers of the seed 1234, while being evaluated every 100 updates using seed 4321. Hence, only the first two instances, i.e., N50a and N50b, were known to the agent. Instance N100c has a new customer distribution (seed 3241). By doing so, we aim to test the transferability and generalization of our ASAP-NCO model. As shown in Table 2, the ASAP method achieves better results in all cases examined. The highest relative deviation between applied heuristics and ASAP method was found in the instances N500a and N1000a. We suppose that this is due to the fact that in the prize-collecting CVRP with a finite fleet, the OR heuristics prefer the routes with a higher number of customers and seek a trade-off between maximizing visited nodes and minimizing distance. The goal of both the learning agent and the heuristics was to collect the maximum demand (customers, in the case of OR heuristics), constrained by the number of available vehicles, while minimizing the total mileage. However, the ASAP agent is rewarded for maximizing the demand collected, as described in Section 3.2. Thus, the ASAP agent pays more attention to the nodes with high demand resulting in fewer customers, whereas the heuristics attempts to maximize the number of customers in a tour resulting in longer tours. For small problems, where the number of possible optima is limited, the heuristics and NCO solutions are of the same magnitude. However, the deviation in favor of ASAP method becomes particularly evident as the number of possible alternative solutions grows with the size of the problem instance. Remarkably, in the N1000c instance with a new customer distribution, the ASAP solution was still 25% better than the benchmarked heuristics. This highlights the ability of the presented approach to generalize, even when the instance has much more nodes than those it was trained on and an unknown distribution. Note that we have only measured the total distance traveled as a solution metric and do not include, for example, prizes or rewards for visiting customers when validating the results. Figure 4 visualizes a comparison of routes generated by PyVRP and the ASAP method for two instances, N50a and N1000c. In smaller instances, such as N50a, the routes constructed by the ASAP agent are comparable to those produced by the heuristic solutions. There are only slight overlaps, probably due to time constraints. In the N1000c instance, the ASAP method results in a significantly lower number of customers per tour compared to the benchmarked heuristics. This supports the hypothesis that the ASAP agent selects nodes with higher demand, leading to shorter tours, while the compared heuristics target nodes with more uniform demand. Of course, we ensured that the total collected demand was maximized in both cases for the given vehicle fleet. The policy training of the ASAP-NCO model took us approximately one week on the CPU cluster machine.

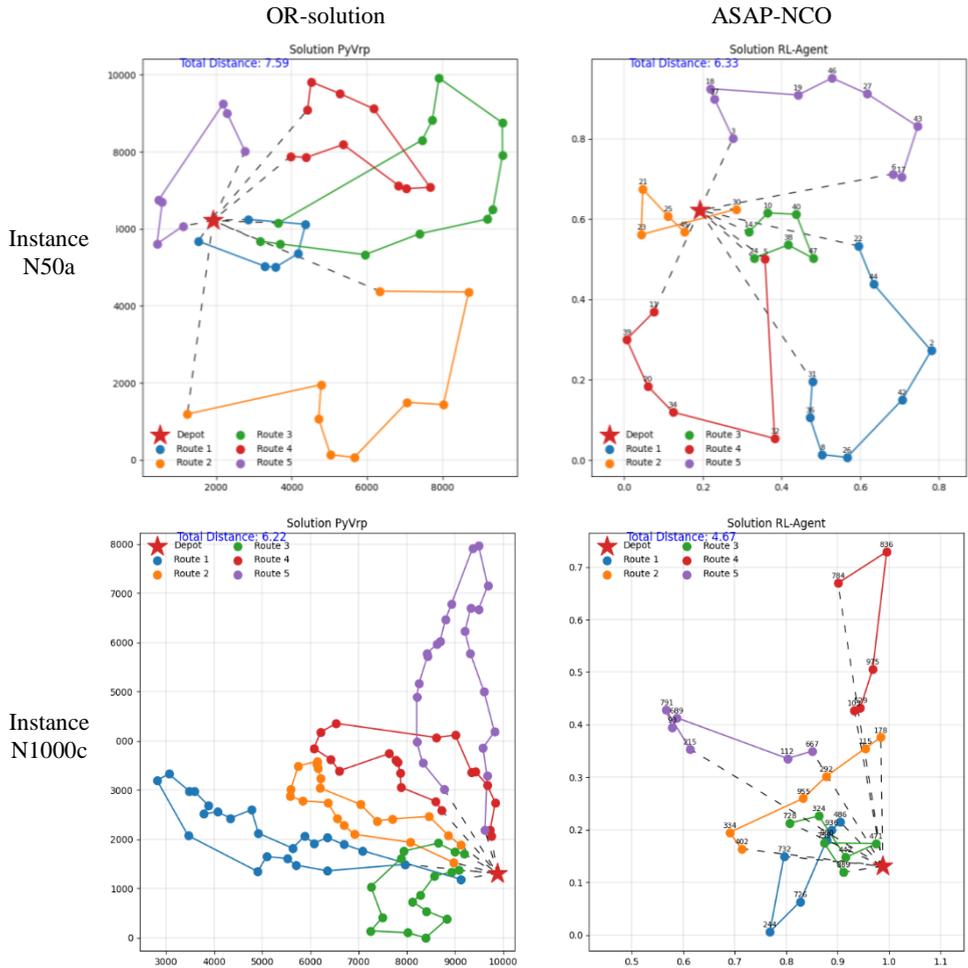

Figure 4: Instance 50a (top row) and 1000c (bottom row). The OR solutions (left column) and the corresponding ASAP solutions (right column) for selected instances. For the instance N50a, the OR heuristics result in total costs of 7.59 distance units (top-left), compared to the 6.33 distance units from the ASAP method (top-right). In the large instance N1000c, the OR solution results in 6.22 distance units (bottom left), while ASAP-NCO generates 5 tours of 4.67 distance units in total (bottom-right).

## 5. Conclusion and outlook

In this study, an end-to-end NCO model for a capacitated VRP with time constraints and a finite vehicle fleet was presented—an NP-hard, multi-objective combinatorial problem often encountered in the context of transport logistics. Here, we adopted the encoder-decoder policy, trained by the Proximal Policy Optimization (PPO) algorithm (Huang et al. 2022). We adopted the Policy Optimization with Multiple optima (POMO) protocol based on Kwon et al. (2020) following the implementation scheme proposed by Wan et al. (2023), which we found to be the most efficient and flexible method for training the end-to-end NCO algorithms. An ASAP approach was introduced to handle dynamic customer constraints such as time windows. By rescaling attention scores based on direct environment information, we present an easy-to-follow alternative for implementing dynamic constraints in sophisticated encoder-decoder architectures. Through the amplification of attention scores in the pointer network, our method effectively prioritized customers with urgent services during the tour construction. It can be easily extended to other soft and hard constraints, such as additional service fees, customer preferences, or distance constraints for electric vehicles. Moreover, we successfully designed and adjusted a reward function that penalizes the agent for incomplete trips, achieving multi-objective optimization. The tested model demonstrated reasonable and reliable results and showed good generalization when applied to new problem instances. Although the results of our model significantly outperformed those of the benchmarked heuristics, especially for the large problem instances, we identified that this advantage was primarily due to the mismatch in the objective functions. After the in-depth analysis of the tour compositions presented in Sections 4.2 and 4.3, we demonstrated that the solutions produced by our method are both adequate and still optimal. However, validating the optimality of solutions generated by NCO is a non-trivial task. We encourage interested readers to examine the NCO-produced solutions, especially in scenarios involving multi-objective, multi-constrained real-world transportation problems.

NCO holds exciting prospects for the future of optimization, particularly in transport optimization. In our future research, we aim to further develop and implement our method within a real-world case study, focusing on more integrated problems such as multi-depot and heterogeneous fleet optimization.

**References**

Ahmed, Z. H., & Yousefikhoshbakht, M. (2023). A Hybrid Algorithm for the Heterogeneous Fixed Fleet Open Vehicle Routing Problem with Time Windows. Symmetry, 15(2), 486.

Balas, E. (1989). The prize collecting traveling salesman problem. Networks, 19(6), 621-636.

Barua, L., Zou, B., & Zhou, Y. (2020). Machine learning for international freight transportation management: a comprehensive review. Research in Transportation Business & Management, 34, 100453.

Bello, I., Pham, H., Le, Q. V., Norouzi, M., & Bengio, S. (2016). Neural combinatorial optimization with reinforcement learning. arXiv preprint arXiv:1611.09940.

Berto, F., Hua, C., Park, J., Kim, M., Kim, H., Son, J., ... & Park, J. (2023). Rl4co: an extensive reinforcement learning for combinatorial optimization benchmark. arXiv preprint arXiv:2306.17100.

Bogyrbayeva, A., Meraliyev, M., Mustakhov, T., & Dauletbayev, B. (2022). Learning to solve vehicle routing problems: A survey. arXiv preprint arXiv:2205.02453.

Bräysy, O., & Gendreau, M. (2005). Vehicle routing problem with time windows, Part II: Metaheuristics. Transportation science, 39(1), 119-139.

Brockman, G., Cheung, V., Pettersson, L., Schneider, J., Schulman, J., Tang, J., & Zaremba, W. (2016). Openai gym. arXiv preprint arXiv:1606.01540.

Chen, J., & Luo, J. (2023, September). Enhancing Vehicle Routing Solutions Through Attention-Based Deep Reinforcement Learning. In 2023 5th International Conference on Data-driven Optimization of Complex Systems (DOCS) (pp. 1-9). IEEE.

Da Costa, P. R. D. O., Rhuggenaath, J., Zhang, Y., & Akcay, A. (2020). Learning 2-opt heuristics for the traveling salesman problem via deep reinforcement learning. arXiv preprint arXiv:2004.01608.

Deudon, M., Cournut, P., Lacoste, A., Adulyasak, Y., & Rousseau, L. M. (2018). Learning heuristics for the tsp by policy gradient. In Integration of Constraint Programming, Artificial Intelligence, and Operations Research: 15th International Conference, CPAIOR 2018, Delft, The Netherlands, June 26–29, 2018, Proceedings 15 (pp. 170-181). Springer International Publishing.

Eshragh, A., Esmaeilbeigi, R., & Middleton, R. (2020). An analytical bound on the fleet size in vehicle routing problems: A dynamic programming approach. Operations Research Letters, 48(3), 350-355.

Falkner, J. K., & Schmidt-Thieme, L. (2020). Learning to solve vehicle routing problems with time windows through joint attention. arXiv preprint arXiv:2006.09100.

Fitzpatrick, J., Ajwani, D., & Carroll, P. A Scalable Learning Approach for the Capacitated Vehicle Routing Problem. Available at SSRN 4633199.

Gong, Y. J., Zhang, J., Liu, O., Huang, R. Z., Chung, H. S. H., & Shi, Y. H. (2011). Optimizing the vehicle routing problem with time windows: A discrete particle swarm optimization approach. IEEE Transactions on Systems, Man, and Cybernetics, Part C (Applications and Reviews), 42(2), 254-267.

Goodson, J. C., Ohlmann, J. W., & Thomas, B. W. (2013). Rollout policies for dynamic solutions to the multivehicle routing problem with stochastic demand and duration limits. Operations Research, 61(1), 138-154.

Heggen, H., Molenbruch, Y., Caris, A., & Braekers, K. (2019). Intermodal container routing: integrating long-haul routing and local drayage decisions. Sustainability, 11(6), 1634.

Homberger, J., & Gehring, H. (1999). Two evolutionary metaheuristics for the vehicle routing problem with time windows. INFOR: Information Systems and Operational Research, 37(3), 297-318.

Hottung, A., & Tierney, K. (2019). Neural large neighborhood search for the capacitated vehicle routing problem. arXiv preprint arXiv:1911.09539.

Huang, S., Dossa, R. F. J., Raffin, A., Kanervisto, A., & Wang, W. (202b). The 37 implementation details of proximal policy optimization. The ICLR Blog Track 2023.

Huang, S., Dossa, R. F. J., Ye, C., Braga, J., Chakraborty, D., Mehta, K., & Araújo, J. G. (2022a). Cleanrl: High-quality single-file implementations of deep reinforcement learning algorithms. Journal of Machine Learning Research, 23(274), 1-18.

Joshi, C. K., Laurent, T., & Bresson, X. (2019). An efficient graph convolutional network technique for the travelling salesman problem. arXiv preprint arXiv:1906.01227.

Jungwirth, A., Frey, M., & Kolisch, R. (2020). The vehicle routing problem with time windows, flexible service locations and time-dependent location capacity.

Kool, W., Van Hoof, H., & Welling, M. (2018). Attention, learn to solve routing problems!. arXiv preprint arXiv:1803.08475.

Kritzinger, S., Tricoire, F., Doerner, K. F., Hartl, R. F., & Stützle, T. (2017). A unified framework for routing problems with a fixed fleet size. International Journal of Metaheuristics, 6(3), 160-209.

Kwon, Y. D., Choo, J., Kim, B., Yoon, I., Gwon, Y., & Min, S. (2020). Pomo: Policy optimization with multiple optima for reinforcement learning. Advances in Neural Information Processing Systems, 33, 21188-21198.

Li, Zhang, Wang, Wang, Han (2021a). Deep Reinforcement Learning for Combinatorial Optimization: Covering Salesman Problems. Covering Salesman Problem.

Li, J., Ma, Y., Gao, R., Cao, Z., Lim, A., Song, W., & Zhang, J. (2021b). Deep reinforcement learning for solving the heterogeneous capacitated vehicle routing problem. IEEE Transactions on Cybernetics, 52(12), 13572-13585.

Liang, E., Liaw, R., Nishihara, R., Moritz, P., Fox, R., Goldberg, K., ... & Stoica, I. (2018, July). RLlib: Abstractions for distributed reinforcement learning. In International conference on machine learning (pp. 3053-3062). PMLR.

Lu, H., Zhang, X., & Yang, S. (2020). A learning-based iterative method for solving vehicle routing problems. In International conference on learning representations.

Nazari, M., Oroojlooy, A., Snyder, L., & Takác, M. (2018). Reinforcement learning for solving the vehicle routing problem. Advances in neural information processing systems, 31.

Novoa, C., & Storer, R. (2009). An approximate dynamic programming approach for the vehicle routing problem with stochastic demands. European Journal of Operational Research, 196(2), 509–515. doi:10.1016/j.ejor.2008.03.023

Peng, B., Wang, J., & Zhang, Z. (2020). A deep reinforcement learning algorithm using dynamic attention model for vehicle routing problems. In Artificial Intelligence Algorithms and Applications: 11th International Symposium, ISICA 2019, Guangzhou, China, November 16–17, 2019, Revised Selected Papers 11 (pp. 636-650). Springer Singapore.

Powell, W. B. (2019). A unified framework for stochastic optimization. European Journal of Operational Research, 275(3), 795-821.

Powell, W. B. (2021). Reinforcement learning and stochastic optimization. https://castle.princeton.edu/wp-content/uploads/2021/06/RLSO-Cover-Chapter9-June302021.pdf [last accessed on 06.2023]

Rabecq, B., & Chevrier, R. (2022). A deep learning Attention model to solve the Vehicle Routing Problem and the Pick-up and Delivery Problem with Time Windows. arXiv preprint arXiv:2212.10399.

Schulman, J., Wolski, F., Dhariwal, P., Radford, A., & Klimov, O. (2017). Proximal policy optimization algorithms. arXiv preprint arXiv:1707.06347.

Simao, H. P., Day, J., George, A. P., Gifford, T., Nienow, J., & Powell, W. B. (2009). An approximate dynamic programming algorithm for large-scale fleet management: A case application. Transportation Science, 43(2), 178-197.

Solozabal, R., Ceberio, J., & Takáč, M. (2020). Constrained combinatorial optimization with reinforcement learning. arXiv preprint arXiv:2006.11984.

Stenger, A., Schneider, M., & Goeke, D. (2013). The prize-collecting vehicle routing problem with single and multiple depots and non-linear cost. EURO Journal on Transportation and Logistics, 2(1-2), 57-87.

Stimpson, D., & Ganesan, R. (2015). A reinforcement learning approach to convoy scheduling on a contested transportation network. Optimization Letters, 9, 1641-1657.

Tessler, C., Mankowitz, D. J., & Mannor, S. (2018). Reward constrained policy optimization. arXiv preprint arXiv:1805.11074.

Vaswani, A., Shazeer, N.M., Parmar, N., Uszkoreit, J., Jones, L., Gomez, A.N., Kaiser, L., & Polosukhin, I. (2017). Attention is All you Need. ArXiv, abs/1706.03762.

Vinyals, O., Fortunato, M., & Jaitly, N. (2015). Pointer networks. Advances in neural information processing systems, 28.

Wan, C. P., Li, T., & Wang, J. M. (2023). RLOR: A Flexible Framework of Deep Reinforcement Learning for Operation Research. arXiv preprint arXiv:2303.13117.

Wang, Q., & Tang, C. (2021). Deep reinforcement learning for transportation network combinatorial optimization: A survey. Knowledge-Based Systems, 233, 107526.

Wang, Y., Jia, Y. H., Chen, W. N., & Mei, Y. (2024). Distance-aware Attention Reshaping: Enhance Generalization of Neural Solver for Large-scale Vehicle Routing Problems. arXiv preprint arXiv:2401.06979.

Wouda, N. A., Lan, L., & Kool, W. (2024). PyVRP: A high-performance VRP solver package. INFORMS Journal on Computing.

Xin, L., Song, W., Cao, Z., & Zhang, J. (2021, May). Multi-decoder attention model with embedding glimpse for solving vehicle routing problems. In Proceedings of the AAAI Conference on Artificial Intelligence (Vol. 35, No. 13, pp. 12042-12049).